\documentclass[sigconf, nonatbib]{acmart}

\AtBeginDocument{%
  }

\setcopyright{acmlicensed}
\copyrightyear{2024}
\acmYear{2024}
\acmDOI{XXXXXXX.XXXXXXX}

\acmConference[ACM BCB '24]{The 15th ACM Conference on Bioinformatics, Computational Biology, and Health Informatics}{Nov. 22-25, 2024}{Shenzhen, Guangdong, PR China}

\usepackage{color}
\usepackage{setspace}
\usepackage{booktabs}
\usepackage{todonotes}
\usepackage{multicol}
\usepackage{longtable}
\usepackage{tabularray}
\usepackage{caption,subcaption}

\usepackage{upgreek}

\usepackage{url}
\usepackage{multirow}
\usepackage{setspace}
\usepackage{algpseudocode}
\usepackage{algorithm2e}
\LinesNumbered
\RestyleAlgo{algoruled}

\usepackage{url}            
\usepackage{amsfonts}       
\usepackage{nicefrac}       
\usepackage{microtype}      
\usepackage{ulem}
\usepackage{caption}
\usepackage{subcaption}

\usepackage{lineno}
\usepackage{longtable}
\usepackage{lscape}
\usepackage[final]{pdfpages}

\definecolor{DarkRed}{rgb}{0.5,0.1,0.1}

\newcommand{\model}{MixEHR-Nest\xspace}

\usepackage{adjustbox}
\usepackage{upgreek}
\usepackage{bm}

\usepackage{multirow}
\usepackage{longtable}
\usepackage{colortbl}

\usepackage{lineno,hyperref}

\begin{document}


\title[MixEHR-Nest]{MixEHR-Nest: Identifying Subphenotypes within Electronic Health Records through Hierarchical Guided-Topic Modeling}

\author{Ruohan Wang}
\authornote{These authors contributed equally to this research.}
\affiliation{%
  \institution{School of Computer Science, McGill University}
  \city{Montreal}
  \state{Quebec}
  \country{Canada}}

\author{Zilong Wang}
\authornotemark[1]
\orcid{0009-0000-5600-7437}
\affiliation{%
  \institution{Department of Biomedical Engineering, Faculty of Medicine, McGill University}
  \city{Montreal}
  \state{Quebec}
  \country{Canada}}

\author{Ziyang Song}
\affiliation{%
  \institution{School of Computer Science, McGill University}
  \city{Montreal}
  \state{Quebec}
  \country{Canada}}

\author{David Buckeridge}
\authornotemark[1]
\affiliation{%
  \institution{School of Population and Global Health, McGill University, McGill University}
  \city{Montreal}
  \state{Quebec}
  \country{Canada}}

\author{Yue Li}
\authornote{Correspondence to yueli@cs.mcgill.ca}
\affiliation{%
  \institution{School of Computer Science, McGill University}
  \city{Montreal}
  \state{Quebec}
  \country{Canada}}


\renewcommand{\shortauthors}{Wang et al.}

\begin{abstract}
Automatic subphenotyping from electronic health records (EHRs) provides numerous opportunities to understand diseases with unique subgroups and enhance personalized medicine for patients. However, existing machine learning algorithms either focus on specific diseases for better interpretability or produce coarse-grained phenotype topics without considering nuanced disease patterns.
In this study, we propose a guided topic model, \model, to infer subphenotype topics from thousands of disease using multi-modal EHR data. Specifically, \model detects multiple subtopics from each phenotype topic, whose prior is guided by the expert-curated phenotype concepts such as Phenotype Codes (PheCodes) or Clinical Classification Software (CCS) codes.
We evaluated \model on two EHR datasets: (1) the MIMIC-III dataset consisting of over 38 thousand patients from intensive care unit (ICU) from Beth Israel Deaconess Medical Center (BIDMC) in Boston, USA; (2) the healthcare administrative database PopHR, comprising 1.3 million patients from Montreal, Canada.
Experimental results demonstrate that \model can identify subphenotypes with distinct patterns within each phenotype, which are predictive for disease progression and severity. Consequently, \model distinguishes between type 1 and type 2 diabetes by inferring subphenotypes using CCS codes, which do not differentiate these two subtype concepts. Additionally, \model not only improved the prediction accuracy of short-term mortality of ICU patients and initial insulin treatment in diabetic patients but also revealed the contributions of subphenotypes. 
For longitudinal analysis, \model identified subphenotypes of distinct age prevalence under the same phenotypes, such as asthma, leukemia, epilepsy, and depression. The \model software is available at GitHub: \hyperlink{https://github.com/li-lab-mcgill/MixEHR-Nest}{https://github.com/li-lab-mcgill/MixEHR-Nest}.
\end{abstract}

\ccsdesc[500]{Applied computing~Life and medical sciences}
\ccsdesc[500]{Computing methodologies~Machine learning}
\keywords{Electronic health records, Multi-modality, Topic modeling, Subphenotyping, Expert-guidance, Gibbs sampling}

\maketitle

\section{Introduction}
The widespread adoption of electronic health records (EHRs) offers numerous opportunities to refine medical concepts and automate disease diagnosis. EHRs contain a heterogeneous collection of patient health data, comprising structured data such as International Classification of Diseases (ICD) codes, Diagnosis-Related Group (DRG) codes, and medication codes (RxNorm), as well as unstructured data such as clinical notes. Distilling interpretable phenotype representations from multi-modal data can enhance the understanding of patient health status and the prediction of disease onset. In clinical practice, diseases often occur as subphenotypes, which are subgroups of traits within the same disease label but with distinct phenotypic characteristics. These subphenotypes can differ in severity and underlying pathology, thereby explaining varied responses to the same medical treatments and promoting precision medicine research.

Traditional phenotyping methods rely on human-curated phenotype concepts, such as Phenotype Codes (PheCodes) and Clinical Classification Software (CCS) codes, which group ICD diagnostic codes into clinically relevant concepts for medical informatics and clinical reasoning research ~\cite{wei2015extracting}.
However, these expert-defined approaches often fail to capture the nuanced diversity of diseases, resulting in inflexible representations that lack fine-grained phenotypic patterns. Machine learning algorithms offer a promising direction for incorporating nuanced patterns into phenotype representations. Various unsupervised machine learning methods have been developed to derive clinically meaningful phenotype representations from EHR data, aiding in disease risk prediction and personalized treatment recommendation. Despite their potential, these methods often require manual interpretation of unobserved patients or phenotype embeddings, limiting their scalability to produce identifiable clusters from thousands of phenotypes.

In this study, we present \model, a seed-guided hierarchical topic model that distinguishes fine-grained subphenotype topics without compromising interpretability and scalability. To model EHR data, we treat each patient’s medical history, and its codes (i.e. ICD codes) are treated as a document and word tokens, respectively ~\cite{li2020inferring, wen2021mining}. Our study includes three key technical contributions: (1) a seed-guided topic model that discovers multiple subphenotype topics within each phenotype; (2) a multi-modality modeling that learn multiple types of EHR information; (3) automatic subphenotyping scaled up to thousands of phenotypes with high interpretability. Experimental results demonstrate that \model is capable of inferring distinct subphenotype topics from over 1500 phenotypes using two large-scale EHR datasets, i.e., Medical Information Mart for Intensive Care (MIMIC-III) and PopHR datasets. The inference of subphenotype topics enhances diverse healthcare tasks, including ICU mortality prediction, initial insulin recommendation, and longitudinal disease prognosis. 

\section{Related Works}
Existing research on automatic subphenotyping has largely focused on clustering techniques ~\cite{reddy2020subphenotypes, lasko2013computational}, non-negative matrix/tensor factorization ~\cite{ho2014limestone}, mixture models ~\cite{mayhew2018flexible} or autoencoder methods ~\cite{xu2020identifying, baytas2017patient, weng2020deep}. These methods typically struggle with poor identifiability when applied to large, heterogeneous cohort with thousands of diseases. Consequently, most research focus on single diseases groups, such as acute respiratory distress syndrome (ARDS) ~\cite{maddali2022subtype, calfee2014subtype, Sinha2018subtype}, asthma ~\cite{wenzel1999evidence, wenzel2012natureasthma}, sepsis ~\cite{seymour2019derivation, mayhew2018flexible, bhavani2019identifying}, and acute kidney injury (AKI) ~\cite{xu2020identifying,  bhatraju2019identification}. Scaling these methods to identify subphenotypes across broader disease categories remains a challenge. 

Unsupervised topic models are often seen as a potential solution for handling large topic numbers, but they generally do not incorporate expert knowledge into inference process, 
limiting their ability to identify clinically meaningful phenotypes. Recently, expert-guided topic models like MixEHR-Guided and MixEHR-Seed leverage seed-guided topic inference to infer expert-defined phenotype topics ~\cite{ahuja2022mixehr, song2022automatic}. Built on the heterogeneous topic model MixEHR ~\cite{li2020inferring}, these models can learn from multi-modal EHR data and assign distinct phenotype topic distributions for each modality. However, despite their ability to annotate thousands of phenotypes, they fall short in capturing fine-grained phenotypic patterns for subphenotyping.

Our proposed \model can infer multi-modal subphenotype topics from heterogeneous EHR data, addressing the limitations of prior approaches and offering a scalable solution for identifying subphenotypes for  thousands of diseases simultaneously. Given the lack of established methods that address large-scale automatic subphenotyping for multiple diseases, a direct comparison was not feasible. As a result, we focused on demonstrating the effectiveness of our approach through downstream tasks like mortality prediction (Section 6.2) and insulin usage prediction (Section 6.3), which indirectly show that subphenotypes enhance prediction performance compared to general phenotypes.

\section{Methodology}
Here we provide a high-level overview of the methods. \textbf{Section} \ref{sec:background} outlines the data-generative process of MixEHR-Nest. The key difference from the standard Latent Dirichlet Allocation is the introduction of a phenotype-guided subtopic prior. \textbf{Section} \ref{sec:infer_topic} describes the inference algorithm to infer the subtopic assignment (i.e., $z$) of each EHR code. \textbf{Section} \ref{sec:init_stat} details parameter initialization, which is crucial for model performance. Finally, \textbf{Section} \ref{sec:new_patient_phenotype_prediction} describes the inference of disease subtopic mixtures of test patients. All the notations are either defined in \textbf{Table} \ref{table: var_description} or in-text.

\subsection{MixEHR-Nest model generative process}\label{sec:background}
\begin{figure*}[ht!]
\centering
\includegraphics[width = 0.8\linewidth]{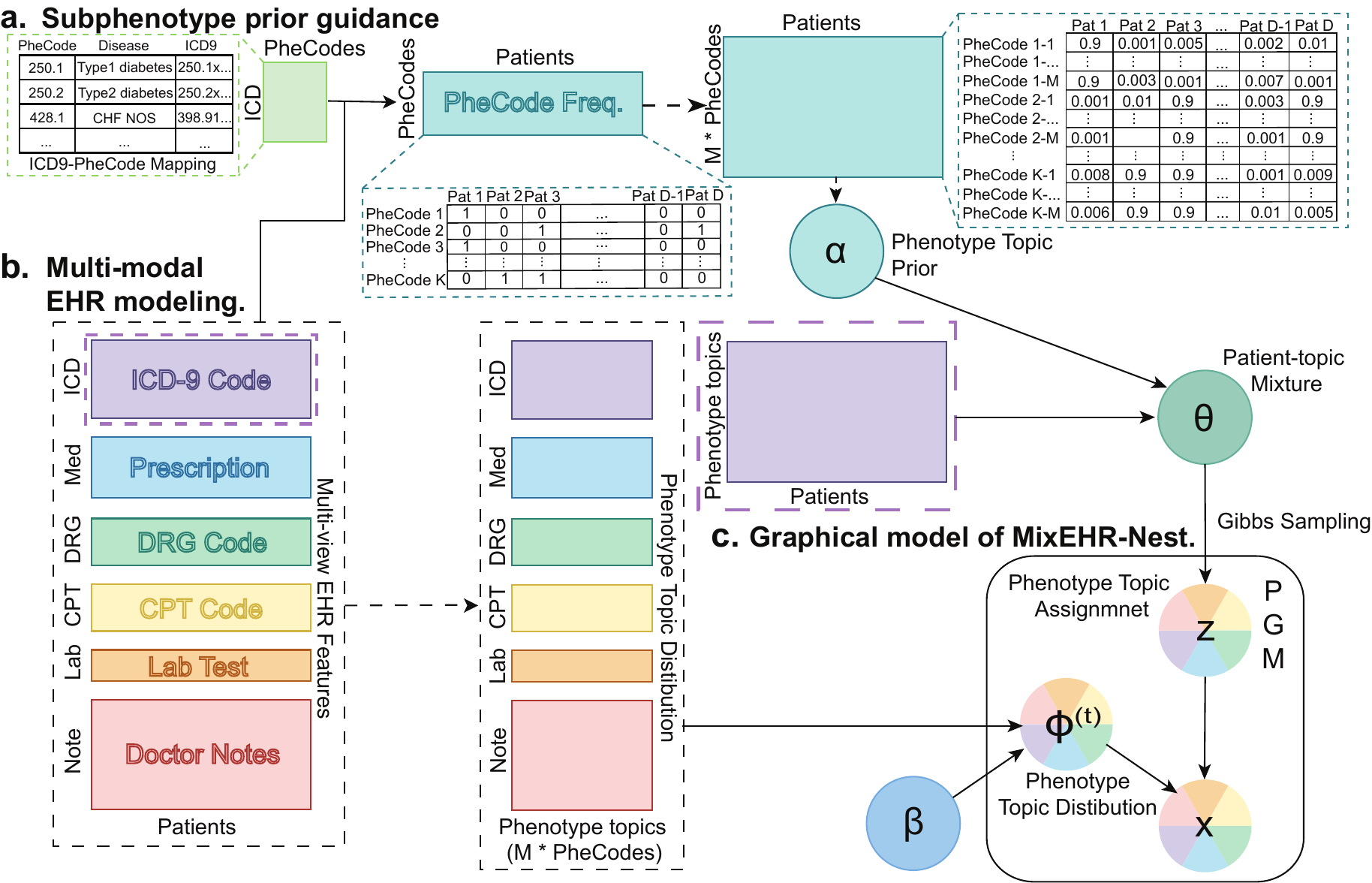}
\vspace{-0.2cm}
\caption{
  \textbf{Schematic of \model on the MIMIC EHR data.}
(a) Subphenotype prior guidance. For each patient d, \model initializes its phenotype topic prior $\alpha_{d,k_m}$ for the subtopic $m$ of the phenotype topic $k$ by computing PheCode occurrence.
(b) Multi-modal EHR modeling. \model learns multi-modal phenotype topics $\bm{\upphi}^{(t)}$ for the  modality $t$. 
(c) Graphical model of \model. The topic mixture $\bm{\uptheta}_{d}$ is drawn from the Dirichlet distribution with $\bm{\upalpha}_d$. For an EHR token $i$ from the modality $t$, the topic assignment $z_{id}$ is sampled from a categorical distribution with $\bm{\uptheta}_d$. Given the topic assignment $z_{id}$ = ${k_m}$, the EHR token $x_{id}$ is then sampled from a categorical distribution with $\bm{\upphi}^{(t)}_{z_{id}}$.
}
\label{fig_model}
\vspace{-0.3cm}
\end{figure*}

In this study, \model takes into account the subphenotypes by assigning $M$ subtopics within each of $K$ phenotype topics, resulting in a total of $K \times M$ topics. For each patient $d$ out of $D$ patients, its topic mixture $\bm{\uptheta}_d$ follows a Dirichlet distribution with  asymmetric topic priors $\boldsymbol{\alpha}_d \in R^{K\times M}$, which
guide posterior topic inference for subphenotypes (Fig.\ref{fig_model} a). For each EHR code $i$ from the patient $d$, \model infers a topic assignment $z_{id}$ sampled from a Categorical distribution based on the topic mixture $\bm{\uptheta}_d$. Its EHR code $x_{id}$ is sampled from the Dirichlet-distributed phenotype topic distribution ${\upphi}_{k_m = z_{id}}$ given the topic assignment $z_{id}$.

\begin{table}
\caption{Notations in MixEHR-Nest}
\vspace{-0.3cm}
\label{table: var_description}
\begin{center}
\begin{tabular}{ l | p{5cm} }
\hline
\textbf{\normalsize Notations}  & \textbf{\normalsize Descriptions}  \\
\hline 
$D$ & total number of EHR documents in dataset \\
$N_d$ & total number of features for EHR document $d$ \\
$K$ &  number of phenotype topics \\
$M$ & number of subphenotype under each topics \\
$W^t$ & feature vocabulary for modality $t$ in dataset\\
\hline 
$w \in R^{W^{(t)}}$ & feature index in modality $t$ \\ 
$x^{(t)}_{id} = w$ & feature index in modality $t$ of token $i$ in EHR document $d$ \\
\textbf{$z^{(t)}_{id}$} & topic assignment for feature $x^{(t)}_{id}$ \\ 
$n_{w.k_m}$ &  counts of feature $w$ assigned to subtopic $m$ of phenotype $k$ across all documents \\
$n_{.dk_m}^{(t= \text{ICD})}$ & count of ICD features of document $d$ assigned to subtopic $m$ of phenotype $k$\\
$n_{.dk_m}^{(t= \text{non-ICD})}$ & count of non-ICD features of document $d$ assigned to subtopic $m$ of phenotype $k$\\
\textbf{$\bm{\upalpha} \in R^{D \times K \times M}$} & Dirichlet document-topic priors \\
\textbf{$\beta$} & Dirichlet topic-feature priors\\
\textbf{$\bm{\uptheta} \in R^{D \times K \times M}$}& topic mixture memberships \\ 
\textbf{$\bm{\upphi^{(t)}} \in R^{K \times M \times W^t}$}& topic distribution of modality $t$ \\
\textbf{$\eta$} &  downscale rate for non-ICD features \\
\toprule
\end{tabular}
\end{center}
\vspace{-0.5cm}
\end{table}

To incorporate expert knowledge, we utilize two phenotype taxonomies: (1) 1641 expert-curated PheCodes from the the Phenome-wide association studies (PheWAS) ~\cite{denny2010phewas}; (2) the 281 coarse-grained CCS codes from CCS mapping. Both mapping systems group clinically relevant ICD codes into broadly defined phenotype concepts. Consequently, the model infers each  phenotype topic for either a PheCode or CCS code

The data generative process of MixEHR-Nest is as follows:
\begin{enumerate}
    \item For the subtopic $m \in \{1, \ldots,M \}$ within the disease topic $k \in \{1, \ldots, K\}$ given the EHR modality $t \in \{1,\ldots, T\}$:
    \begin{enumerate}
        \item Sample $K \times M$ global phenotype topics from Dirichlet distribution ${\upphi}_{k_m}^{\left( t\right)} \sim \text{Dir}\left(\boldsymbol{\upbeta}\right)$ 
    \end{enumerate}
    
    \item For each patient  $d \in \{1, \ldots, D\}$, sample a patient-topic mixture $\bm{\uptheta}_{d}\sim\text{Dir} \left(\bm{\upalpha}_d\right)$:    
    \begin{enumerate}
        \item For each EHR token $i \in \{1,\ldots,N^{(t)}_d\}$ among the $N^{(t)}_d$ tokens with modality $t$ for patient $d$:
        \begin{enumerate}
            \item Sample a topic assignment: $z_{id}^{\left( t\right) }\sim\text{Cat} \left(\boldsymbol{\uptheta}_{d}\right).$ 
            \item Sample an EHR code: $x_{id}^{\left( t\right) }\sim\text{Cat}\left(\boldsymbol{\upphi}_{z^{(t)}_{id}}^{\left( t\right) }\right).$
        \end{enumerate}
    \end{enumerate}
\end{enumerate}
where $Dir(.)$ and $Cat(.)$ indicate Dirichlet and Categorical distributions, respectively. To run \model model, we provide three key steps as shown in Fig.\ref{fig_model}. Initially, it calculates the prior probabilities for each subphenotype based on the mapping of ICD codes to PheCodes (CSS codes) for each patient  . We then initialize the sufficient statistics for the topic inference. Finally, we train \model on the multi-modal EHR dataset, where the patient-topic Dirichlet hyperparameters are adjusted from the probabilities computed in the first step. The training process involves detecting subphenotypes from a patient’s EHR data by inferring the posterior distributions of guided subphenotypes under the reference phenotypes due to constrained topic priors. 

\subsection{Inference}\label{sec:infer_topic}
In this section, we provide the equations for the inference of the phenotype-guided prior and modality-specific topic assignments. We detailed the derivation of the collapsed Gibbs sampling of the LDA in \textbf{Appendix} \ref{sec:lda_derive}.

\paragraph{ICD-specific topic inference} 
For each ICD code \( i \) of the patient \( d \), we infer the topic assignment \( z^{(\text{ICD})}_{id} \) indicating the underlying subphenotype topic, given all the other topic assignments $z^{(\text{ICD})}_{\backslash id}$: 

\begin{align}\label{eq:z_icd}
    & z^{(\text{ICD})}_{id} \vert z^{(\text{ICD})}_{\backslash id} \sim \prod_{k_m=1}^{K\times M} 
    \left(\alpha_{dk_m}+n_{\cdot dk_m}^{-\left(id\right)}\right) \left(\frac{\beta+n_{x_{id}\cdot k_m}^{-(id)}}{W_{\text{ICD}}\beta + \sum_{w=1}^{W_{\text{ICD}}} n_{w\cdot k_m}^{-(id)}}\right) 
\end{align}    
The sufficient statistics $n_{\cdot dk_m}^{(\text{ICD})}$ and $n^{(\text{ICD})}_{w\cdot k_m}$ for ICD modality can be updated as follows, where $n^{(ICD)}_{\cdot dk_m}$ represents the total count of tokens assigned to topic $k_m$ for patient $d$ for ICD, and $n^{(ICD)}_{w\cdot k_m}$ represents the count of tokens with value $w$ assigned to topic $k_m$ for ICD across all patients:

\begin{align}
    &n_{\cdot dk_m}^{(\text{ICD})} = \sum_{i=1}^{N^{(\text{ICD})}_d} [z^{(\text{ICD})}_{id} = k_m] \nonumber \\ 
    & n^{(\text{ICD})}_{w\cdot k_m} = \sum_{d=1}^D \sum_{i=1}^{N^{(\text{ICD})}_d}
    [z^{(\text{ICD})}_{id}=k_m][x^{(\text{ICD})}_{id} = w] 
\end{align}

\paragraph{Non-ICD-modality topic inference} 
For the unguided non-ICD modalities, the updates of topic assignments are defined as follows:
\begin{align}
    & z^{(\text{non-ICD})}_{i'd}\sim \prod_{k_m=1}^{K\times M} 
    \left(\alpha_{dk_m} + n_{\cdot dk_m}^{-\left(i', d\right)}\right) \left(\frac{\beta+n_{x_{i' d} \cdot k_m}^{-(i', d)}}{ W_{\text{non-ICD}}\beta+\sum_{w=1}^{W_{\text{non-ICD}}}n_{w \cdot k_m}^{-(i', d)}}\right)
\end{align}
where $n_{\cdot dk_m} = \hat{n}_{.dk_m}^{(\text{ICD})} + \eta \times n_{.dk_m}^{(\text{non-ICD})}\label{eq:n_dk}$ indicate the updated weighted aggregate and $\hat{n}_{.dk_m}^{(\text{ICD})}$ is the updated ICD-derived topic count. This procedure leverages $\hat{n}_{.dk_m}^{(\text{ICD})}$ from ICD-modality to guide the inference of topic assignments from the non-ICD modalities. We can update as follows:
\begin{align}
    & n_{.dk_m}^{(\text{non-ICD})} = \sum_{i'=1}^{N_d^{(\text{non-ICD})}} [z^{(\text{non-ICD})}_{i'd} = k_m] \nonumber \\
    & n^{(\text{non-ICD})}_{w\cdot k_m} = \sum_{d=1}^D \sum_{i'=1}^{N_d^{(\text{non-ICD})}} [z^{(\text{non-ICD})}_{i'd}=k_m][x^{(\text{non-ICD})}_{i'd}=w]
\end{align}
This is summarized in \textbf{Appendix} \ref{sec:non_icd_topic_infer}.
\paragraph{Cross-sectional topic inference}
For cross-sectional EHR data such as MIMIC-III, we rely on the phenotype-guided topic assignments from the ICD modality as it defines the phenotype concept (e.g., PheCode). We first infer topics on ICD modality iteratively until convergence.  Given the ICD-modality topic assignments, we infer non-ICD modalities iteratively until convergence. For each modality $t$, the convergence is evaluated by the log marginal likelihood:
    $\mathcal{L}^{(t)} =\sum^D_{d=1} \sum^{N^{(t)}_{d}}_{i=1}\sum^{K\times M}_{k_m=1}\log\hat{\uptheta}^{(t)}_{dk_m} \hat{\upphi}^{(t)}_{x^{(t)}_{id}k_m}$ 
where 
    $\hat{\upphi}^{(t)}_{wk_m} \equiv \mathbb{E}\left[\upphi^{(t)}_{wk_m} \mid z_{k_m}\right]=\frac{\beta+n^{(t)}_{w \cdot k_m}}{W^{(t)} \beta+\sum^{W^{(t)}}_{w^{\prime}=1} n^{(t)}_{w^{\prime} \cdot k_m}}$ and 
    $
    \hat{\uptheta}^{(t)}_{dk_m} \equiv \mathbb{E}\left[\uptheta^{(t)}_{dk_m} \mid z_{d}\right] = \frac{\alpha_{dk_m} + n^{(t)}_{\cdot d k_m}}{\sum^{K\times M}_{k_m'=1}\alpha_{dk'_m} +  n^{(t)}_{\cdot d k_m'}}$.

\paragraph{Longitudinal topic inference}
For longitudinal administrative healthcare data such as PopHR, each patients could have multiple visits, where the same code can be recorded multiple times across their medical histories. For each patient, we compute the occurrence of EHR codes across medical visits. We then jointly infer  topic assignments from both ICD and non-ICD modalities by alternating between them at each iteration. We iterate Gibbs samplings until the log likelihood across all modalities converges We provide details in \textbf{Appendix} \ref{sec:longitudinal_topic_infer}. 

\subsection{Initialization}\label{sec:init_stat}

\paragraph{Prior Initialization} \label{sec:prior_init}
For cross-sectional EHR data, where each ICD-9 code is observed only once per patient, we set the hyperparameter $\alpha_{dk_m}$ to 0.9 if the subphenotype $m$ associated with topic $k$ is observed for a patient $d$. Otherwise, we randomly sample the value from a range of [0.001, 0.01]. For longitudinal EHR data, where each code can occur multiple times during patient visits, we estimate probabilities using a two-component Poisson and Lognorm mixture model for each PheCode via  maximum a posteriori (MAP) ~\cite{liao2019highthroughput, ahuja2022mixehr, li2024mixehr}. If a patient $d$ has the PheCodes, we set the hyperparameter $\bm{\upalpha}_{d,k} = (\alpha_{d,k_1}, \ldots, \alpha_{d,k_M})$ to the estimated prior values for phenotype $k$; otherwise, the corresponding topic priors are set to 0. We provide  details about the hyperparameter setup for prior initialization in \textbf{Appendix} \ref{sec:init_prior_eg}

\paragraph{Initialization of sufficient statistics}
We initialize sufficient statistics $n^{(t)}_{\cdot dk_m}=\sum_i [z^{(t)}_{id}=k_m]$ and $n^{(t)}_{w\cdot k_m}=\sum_d\sum_i [z^{(t)}_{id}=k_m][x^{(t)}_{id}=w]$ to align with priors $\bm{\upalpha}$, effectively guiding the posterior inference. Here, $n^{(t)}_{\cdot dk_m}$ represents the total count of tokens assigned to topic $k_m$ for patient $d$ for  modality $t$, and $n^{(t)}_{w\cdot k_m}$ represents the count of tokens with value $w$ assigned to topic $k_m$ for modality $t$ across all patients. We leverage the PheWAS or CCS mapping to project ICD codes to PheCodes or CCS codes.

For ICD modality, for each patient $d$, we examine if its ICD code $x^{(\text{ICD})}_{id}=w$ for $i \in (1,\ldots,N^{(\text{ICD})}_d )$ is associated with the phenotype topic $k$. We then select a random subtopic $m$ within the topic $k$ is selected by incrementing $n_{\cdot dk_m}$ and $n^{(\text{ICD})}_{w\cdot k_j}$ by 1. 

For the tokens from non-ICD modalities, we randomly sample subtopics based on the primary topics observed from the ICD modality (\textbf{Appendix} \ref{sec:eg_init_suff_stat}).

We then compute $n_{\cdot dk_m} = n_{.dk_m}^{(\text{ICD})} + \eta \times n_{.dk_m}^{(\text{non-ICD})}$ by weighting $n^{(t)}_{w\cdot k_m}$ from both ICD and non-ICD modalities, where $\eta$ is a weighting hyperparameter. This aggregation method mitigates the potential impact of inaccurate topic assignments from the non-ICD features, effectively leveraging the guided knowledge from ICD modality. We consider the setup of hyperparameter $\eta \in \{0.2, 0.4, 0.6. 0.8, 1\}$ using a validation set as described in Section \ref{design:death}. 

\subsection{Predict phenotypes for new patients}\label{sec:new_patient_phenotype_prediction}
For a new patient $d'$ with EHR tokens denoted as $x^{(t)}_{id'}$ for $i = 1,\ldots, N^{(t)}_{d'}$, 
we first initialize the patient-topic assignments $n_{.d'k_m}$ the same way as in Section \ref{sec:init_stat}. We then infer topic assignments from ICD and non-ICD modalities, where the non-ICD topic inference leverages the guided information ICD-inferred topic assignments. This is similar to the topic inference during training (Section \ref{sec:infer_topic}) except fixing the estimated topic distributions $\hat{\upphi}^{(t)}_{x^{(t)}_{id'}k_m}$. The detailed algorithm is provided as follows: 
\begin{enumerate}
    \item For each ICD code $i \in (1,\ldots, N^{(ICD)}_{d'})$, we perform Gibbs sampling on $z^{(\text{ICD})}_{id'}$ based on Eq.(\ref{eq:z_icd}) with fixed $\hat{\upphi}^{(\text{ICD})}_{x^{(\text{ICD})}_{id'}k_m}$;

    \item For each non-ICD codes, we perform Gibbs sampling on $z^{(\text{non-ICD})}_{id'}$ as discussed in \textbf{Appendix} \ref{sec:non_icd_topic_infer} with fixed $\hat{\upphi}^{(\text{non-ICD})}_{x^{(\text{non-ICD})}_{id'}k_m}$;

    \item Update expected topic assignment $\hat{\uptheta}_{d'k_m}$ for $d'$ using Eq.(\ref{eq_11});
    \item Evaluate the log marginal likelihood using Eq.(\ref{eq_12});
    \item Repeat 1-4 until the likelihood converges. 
\end{enumerate}
The algorithm is converged when the changes of log marginal likelihood is less than 0.1.  

\section{Dataset and Preprocessing}

\subsection{MIMIC-III dataset}\label{sec:mimic_data}
Medical Information Mart for Intensive Care (MIMIC) III contains de-identified EHR data from over 46,000 patients admitted to the intensive care units (ICU) of Beth Israel Deaconess Medical Center between 2001 and 2012 ~\cite{johnson2016mimic}. The MIMIC-III dataset was obtained and used in accordance with the PhysioNet user agreement.
To preprocess the multi-modal EHR data, we followed the practice described in existing study ~\cite{ahuja2022mixehr}. Specifically, we pre-processed the ICD9 code, current procedures terminology (CPT) code, Diagnosis-Related Group (DRG) code, laboratory tests, medication, and doctor notes. We included the MIMIC dataset of entire admission data and the subjects were divided into single-admission and multiple-admission groups for the analysis in Section \ref{design:death} where the distinction is further explored. 

\subsection{PopHR dataset}\label{sec:pophr_data}
PopHR is a large-scale administrative healthcare data consisting of 1.3 million patients from the Quebec province of Canada between 1998 and 2014 ~\cite{shaban2017pophr}. It includes multiple data sources such as inpatient and outpatient physician claims, hospital discharge abstracts, and outpatient drug claims. We used three data modalities from this dataset: ICD-9 diagnostic codes, administrative procedure codes (ACT), and prescriptions from outpatient physician encounters. For prescription modality, we used active drug ingredients as unique IDs; for the other two modalities, we used ICD-9 and ACT codes, respectively. We removed all features with patient frequencies above 25\% to mitigate the influence of common and non-specific features. The resulting dataset includes 5,738 unique ICD-9 codes, 6,544 ACT codes, and 1,235 drug ingredients for a total of 13,517 unique count features. Additionally, we utilized the PopHR dataset for longitudinal analysis, as the patients can encounter multiple occurrences of ICD-9 code across their medical histories.

\subsection{Phenotype guidance}
We utilized the phenotype concepts from PheWAS and CCS systems with respect to the ICD-9 codes. 
The PheWAS systems maps ICD-9 codes to around 1,800 PheCodes (\url{https://phewascatalog.org/phecodes}) ~\cite{denny2010phewas}. It provides a hierarchical structure of PheCodes, allowing us to capture phenotype concepts from various phenotypic grain. The CCS system maps projects ICD-9 codes to 212 CCS codes ~\cite{hcup_ccs,wei2017evaluating}. In this study, we consider 2-decimal PheCodes and all CCS codes, associated ICD codes are observed in at least one patient in the EHR dataset. Consequently, we obtained 1641 PheCodes and 1570 PheCodes from the MIMIC-III and PopHR datasets, respectively . 

\section{Experiments}
\subsection{ICU mortality prediction using MIMIC-III}\label{design:death}
In this experiment using the MIMIC-III dataset, we evaluate that the modeling of subphenotype topics could improve mortality prediction by analyzing severity levels (Fig.\ref{sup-2} a). We divide the dataset into single and multiple-admission groups (Sec.\ref{sec:mimic_data} and Fig.\ref{sup-2} a). The single-admission group is used to train the MixEHR-Nest model and estimate the topic distribution $\hat{\upphi}$, which is then used to infer the patient-topic mixture $\uptheta$ for the multiple-admission group based on their second-last admission records (Sec.\ref{sec:new_patient_phenotype_prediction} and Fig.\ref{sup-2} a). We avoid the last admission, which contains mortality-related information. The patient-topic mixtures $\uptheta$ of the multiple-admission group are divided into training, validation, and test sets as inputs to classifiers (Fig.\ref{sup-2}a), following standard machine learning practices, except that the input features are the topic mixtures inferred by MixEHR-Nest. We tuned the hyperparameter $\eta$ (Eq \ref{eq:n_dk}) using five different values of $\eta$ to infer topic mixtures $\uptheta$ for both the training and validation sets, resulting in five pairs of $\uptheta_{train}$ and $\uptheta_{val}$. An elastic net or random forest (RF) classifier using the scikit-learn ~\cite{pedregosa2011scikit} was trained on $\uptheta_{train}$ and $Y_{train}$ to predict mortality using $\uptheta_{val}$.
We used the area under precision-recall curve (AUPRC) as the evaluation metric. The experiment  was repeated for 20 random train-validation splits, and the model with the highest mean AUPRC was selected to predict mortality in the test set (Fig.\ref{sup-2} b). We compared two baselines: RF using $\alpha$ and feature count matrices measured by top K precision. 

\subsection{Explaining mortality by subphenotypes}\label{sec:shap}
In this experiment, we assess the contribution of subphenotypes concerning ICU mortality prediction in theMIMIC-III dataset using a RF classifier. We computed global feature importance scores for all subphenotypes used by the RF and calculated SHapley Additive exPlanations (SHAP) values to explain individual mortality predictions ~\cite{lundberg2017unified,lundberg2020local}. For each subject $d$, we calculated  SHAP values for $\hat{\uptheta}_{d,k_m}$ inferred by \model from their second last admission. For the ascertainment analysis, we computed  SHAP values of all subphenotypes in the top 100 patients with the highest mortality risk. We computed SHAP values using the public \href{https://github.com/shap/shap/blob/master/docs/api.rst}{SHAP package}.

\subsection{Insulin Usage Prediction}
To assess the impacts of subphenotypes ondisease severity, we aimed to identify subphenotypes that can predict insulin treatment six months after the initial diabetic diagnosis, indicating the potential exacerbation of diabetic condition. Following the procedure described by ~\cite{song2021supervised}, we extracted 78,712 diabetic patients with ICD-9 codes 250x from the PopHR dataset. These patients had continuous public drug insurance for hospitalization or medical billing, allowing us to track their medication records. We defined the start of insurance coverage as the date of a patient's first diabetes diagnosis. Patients were included if they had continuous public drug insurance following the diagnosis, defined by either (1) having at least 6 months of uninterrupted insurance or (2) having insurance interruptions of less than 2 months. To avoid misclassification due to unrecorded insulin usage, we excluded patients who used insulin within 6 months of their initial diabetes diagnosis. As a results, 11,382 out of 78,712 diabetic patients were labelled as positive, i.e., starting to use insulin 6 months following initial diagnosis of diabetes; the remaining patients were labeled negative. To ensure label balance  (50\% positive patients, 50\% negative patients), 11,382 negative patients were randomly sampled for the experiment. We used an 80/20 train-test split on the total of 22,764 patients.

\subsection{Estimating age-dependent sub-phenotypes}
In this experiment, we aimed to estimate the relative prevalence of its subphenotypes stratified by age using the PopHR dataset, which represents the general population in Montreal ~\cite{shaban2017pophr}. 
For each patient $d$, we computed its patient-topic mixtures $\uptheta_{dk_m}$ as the estimated risk scores for the subtopic $m$ within the phenotype $k$. 
For a given age $t_{age}$, we identified the set of patients $d \in S_{t_{age}}$ such that $T_{min,d} \leq t_{age} \leq T_{max,d}$, where $T_{min,d}$ and $T_{max,d}$ are the first and last recorded ages of patient $d$ in the PopHR dataset. Here, $S_{t_{age}}$ denotes the set of all patients whose recorded age range includes $t_{age}$. We then estimated the population-level prevalence of subtopic $m$ within the phenotype $k$ at age $t_{age}$ as the mean patient-topic mixture over $S_{t_{age}}$:
    $\hat{\rho}_{k_m,{t_{age}}} = {1 \over |S_{t_{age}}|} \sum_{d\in S_{t_{age}}} \hat{\uptheta}_{dk_m}$.
To estimate the relative prevalence over time, we used a predefined sequence of ages $\gamma$ to calculated the relative prevalence at age $t_{age}$ as
    $\hat{\rho}_{k_m,{t_{age}},rel} = {\hat{\rho}_{k_m,{t_{age}}} \over \sum_{t\in \gamma} \hat{\rho}_{k_m,t}}$.
For the baseline prevalence of each phenotype $k$, we replaced the patient topic mixture $\hat{\uptheta}_{dk_m}$ with the patient phenotype counts $n_{dk_m}$ and then calculated the relative baseline population-level prevalence.

\section{Results} 
\subsection{Subphenotype inference from MIMIC-III}
\begin{figure}[!b]
\centering
\includegraphics[width = 0.45\textwidth]{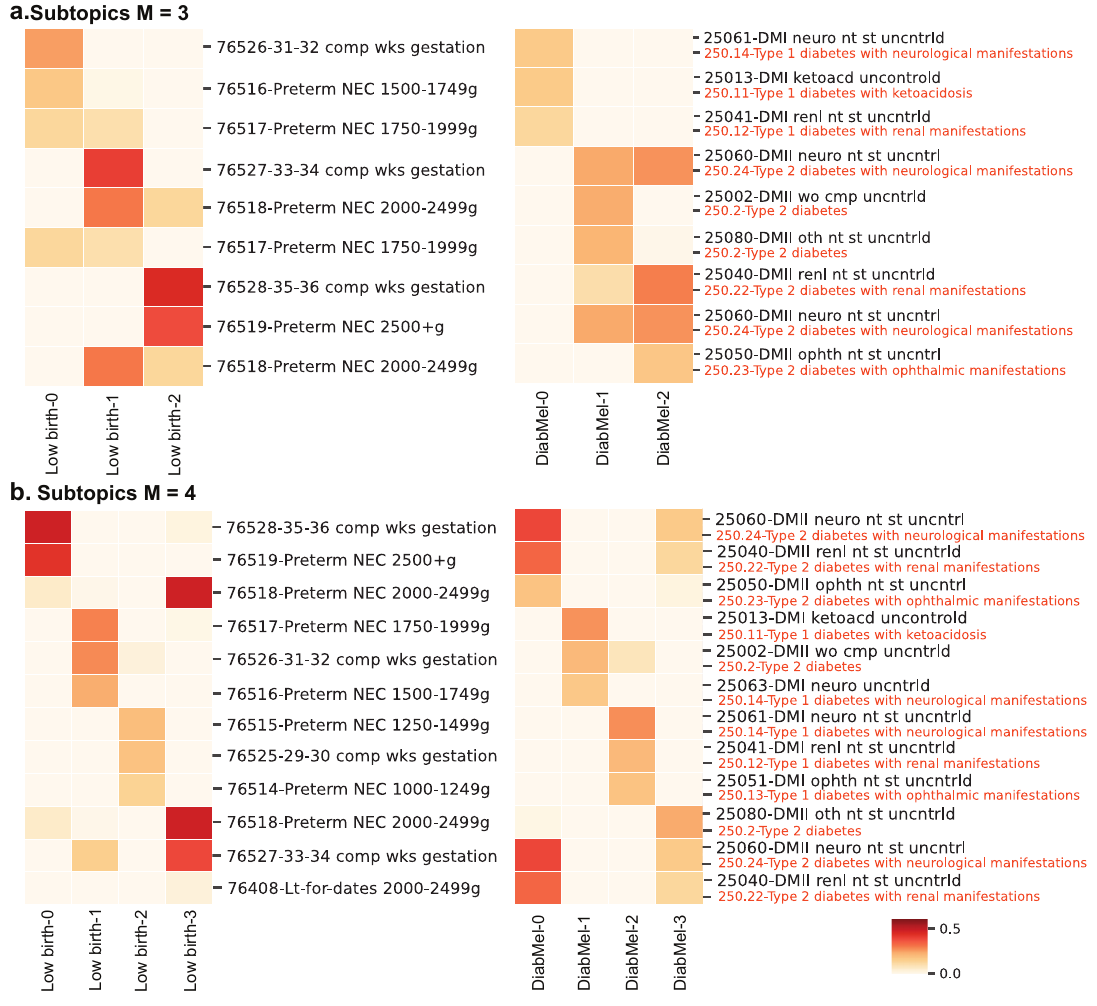}
\vspace{-0.45cm}
\caption{
  \textbf{Top ICD codes inferred by \model for the CCS-guided phenotype topics from the MIMIC-III data for low birth and diabetic phenotypes. As a proof-of-concept, we used PheCode to label ICD9 codes to show that the subtopics of the CCS codes we found reflect the PheCode system, which was not used to train the model.} 
  (a) 3 subtopics per phenotype (M=3). 
  (b) 4 subtopics per phenotype (M=4).
}
\label{fig_subtopic}
\vspace{-0.3cm}
\end{figure}

We first used CCS Codes for phenotype guidance by categorizing ICD-9 Codes into 281 broad disease categories. As a proof-of-concept, we employed the coarse-grained CCS taxonomy to demonstrate that \model can refine these broad categories into clinically meaningful subphenotype topics. Using the MIMIC-III dataset, we showed that \model's subphenotype topic inference using CCS codes '219-Short gestation; low birth weight; and fetal growth retardation' ('low birth') and '50-Diabetes mellitus with complications' ('diabetes'), selected for their detailed granularity levels. This approach highlights the potential of subphenotype topics to enhance disease categorization and improve clinical insights.

To achieve this refinement, selecting the parameter $M$—which determines the number of subphenotype topics—was critical. While in some previous studies, subtypes were not explicitly considered and $M$ was typically set to 1, we explored the benefits of increasing this number. Based on findings in related works, which suggest that values of $M=3$\cite{xu2020subphenotyping} or $M=4$\cite{liu2023subphenotyping} are effective in many cases, we decided to focus on these two settings. 


To demonstrate the impact of subphenotype topic numbers on disease interpretability, we experimented with varying the number $M$ of subtopics per phenotype. At $M=1$, our \model simplified to MixEHR-G, serving as the baseline (Fig.\ref{sup-1}). At $M=3$ and $M=4$, the 'Low birth' subtopics were distinctly delineated by factors such as birth weight and gestational weeks, providing detailed insights into preterm birth (Fig.\ref{fig_subtopic}). Although both levels correspond to PheCode 637.0, $M=4$ provided a more granular categorization, identifying a novel subphenotype for patients with 29-30 weeks gestation and specific weight brackets of 1,000-1,499 grams and 1250-1499 grams.

For 'diabetes' topic, \model at $M=3$ separated Type 1 Diabetes (T1D) with complications (DiabMel-0) from Type 2 Diabetes (T2D), which was further subdivided by complications (DiabMel-1 and DiabMel-2) (Fig.\ref{fig_subtopic} a). At $M=4$, DiabMel-0 was distinctly characterized by severe T1D complications, such as ketoacidosis and neurological issues, highlighting \model's ability to identify acute conditions within T1D. This suggests that for diseases with varying degrees of severity, such as diabetes, an increase in $M$ can uncover more acute subphenotypes.

In this section, the results demonstrated that \model effectively distinguishes subphenotypes, refining from broad classifications at $M=1$ to fine-grained subphenotypes at $M=4$. This granularity, particularly in the nuanced separation of diabetes complications, demonstrates its potential for automating personalized sub-phenotyping.

\subsection{High-risk sub-phenotypes in MIMIC-III}

\begin{figure}[b!]
\centering
\includegraphics[width = 0.8\linewidth]{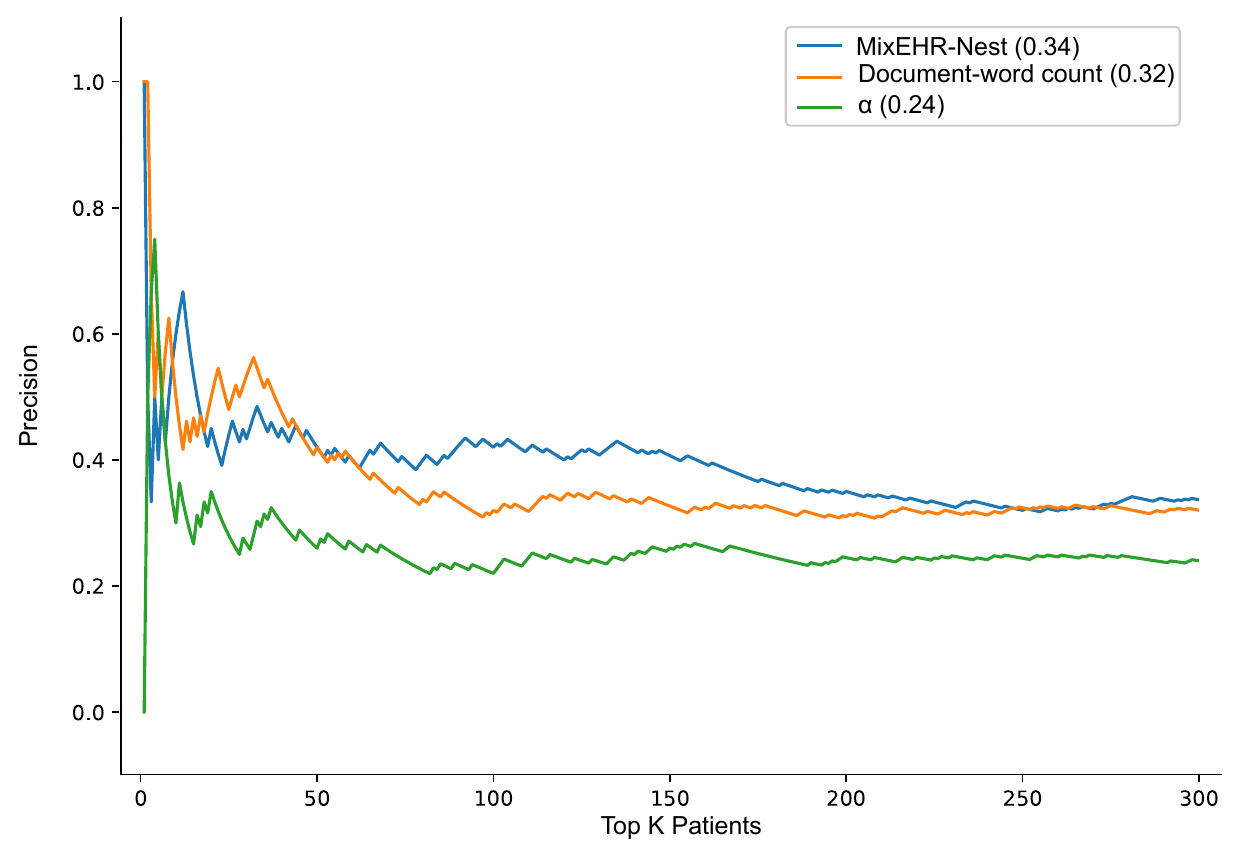}
\vspace{-0.3cm}
\caption{
  \textbf{Top K precision of ICU mortality prediction.}
}
\label{fig_prc}
\end{figure}

We utilized \model to predict patient mortality based on training data from the second-last admission over eight months. As shown in Fig. \ref{fig_prc}, it illustrated the precision of various models for predicting ICU mortality among the top K patients. \model achieved the highest precision of 0.34 for the top 300 patients. All three methods use RF with the inputs including: \model with patient topic mixture $\uptheta_{test}$,  document feature counts ${n_dw}$, and the prior $\alpha_{test}$. Our \model conferred an AUPRC of 33.52\%, which is the highest among these tested methods (Fig.\ref{sup_prc}). 

\begin{figure}[t!]
\centering
\includegraphics[width = \linewidth]{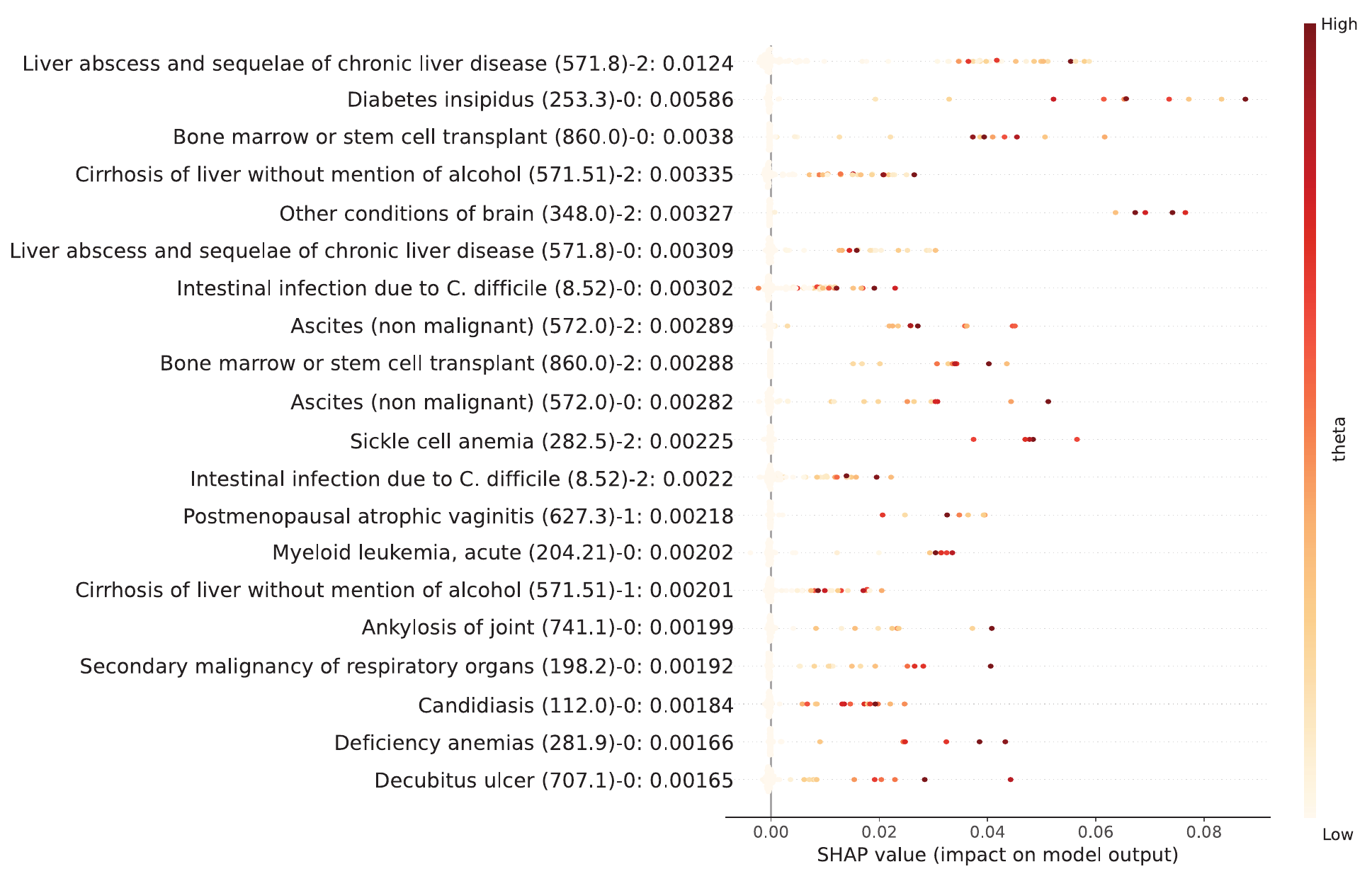}
\vspace{-0.6cm}
\caption{
  \textbf{Analysis of high-risk mortality disease predicted by \model.} SHAP summary plot illustrating the impact of the top 20 high-risk disease subphenotypes on model output for 100 high-risk patients who died in the ICU. Each dot represents the SHAP value for a subphenotype $k_m$ in a sample $d$, with color indicating MixEHR-Nest estimated $\hat{\uptheta}_{dk_m}$.
  }
\label{fig_SHAP_value}
\vspace{-0.3cm}
\end{figure}

To assess subphenotypes predictive of mortality, we used both RF feature importance and SHAP value to evaluate the contribution of individual subphenotypes to ICU mortality prediction (Section \ref{sec:shap}) ~\cite{lundberg2017unified}. The top 100 patients with the highest mortality risk were selected for downstream analysis. SHAP value analysis identified important subphenotypes "Liver abscess and sequelae of chronic liver disease (571.8-2)", "Diabetes Insipidus (253.3-0)", "Bone marrow or stem cell transplant (860-0)", "Cirrhosis of liver without mention of alcohol (571.51-2)", and "Other Conditions of brain (348.0-2)" with the highest mean SHAP values among high-risk patients (Fig.\ref{fig_SHAP_value}). These findings were consistent with the top subphenotypes identified by RF feature importance (Fig.\ref{sup_SHAP_value}). This concordance between SHAP values and RF feature importance, highlighting \model's ability to identify critical predictors of mortality risk.

\begin{figure*}[t!]
\centering
    \includegraphics[width=\linewidth]{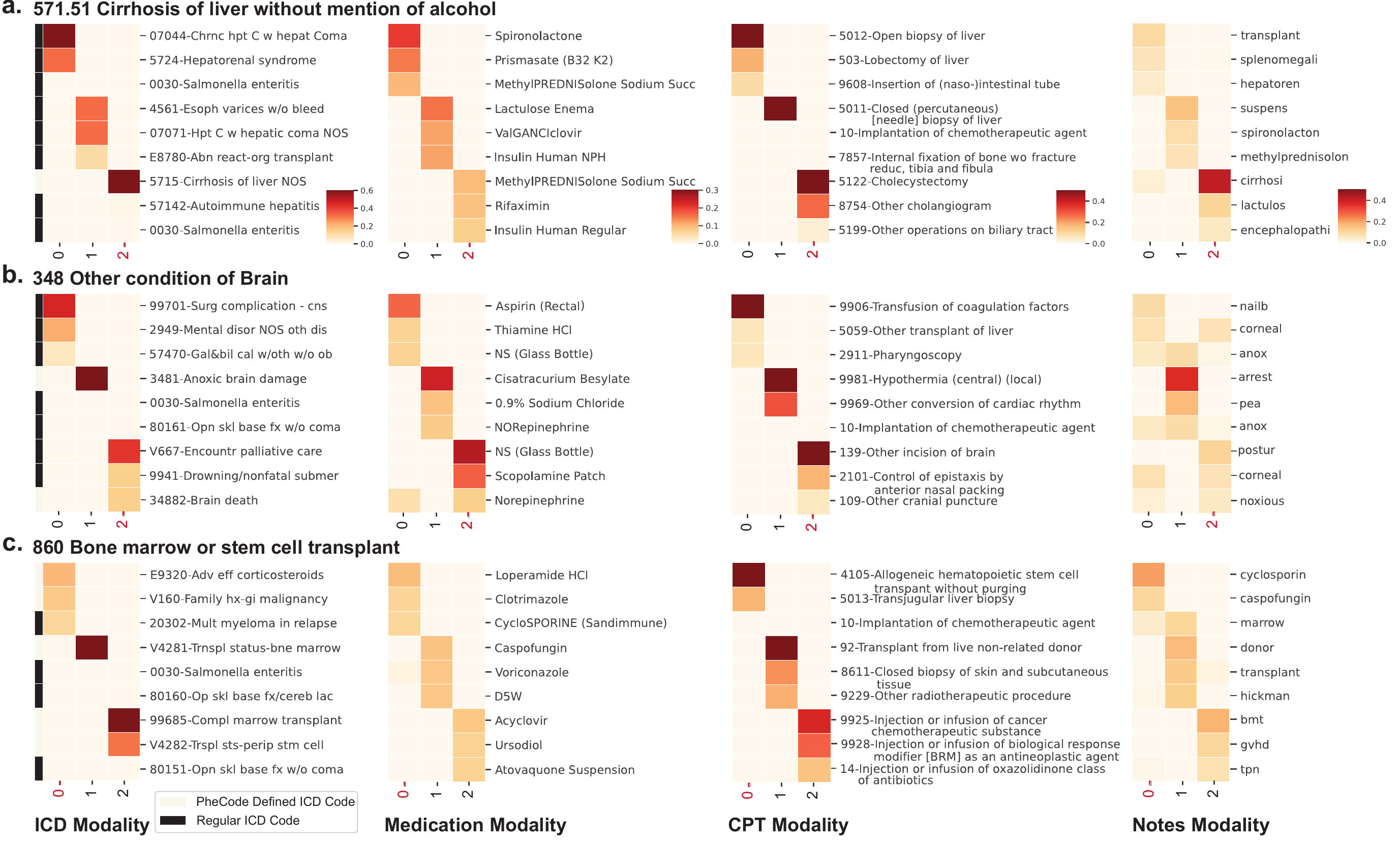}
    \caption{
      \textbf{High-Risk Mortality Phenotype.} The first column indicates the predominant ICD codes for each topic, with sidebars highlighting the PheCode-defining ICD-9 codes. The figure at the second column indicates the primary medications correlated with each topic. The figure at the third column indicates the chief CPT codes associated with each topic. The figure at the last column indicates key feature from doctor notes related to each topic. (a) Cirrhosis of liver without alcohol subphenotypes (571.51). (b) Other conditions of brain (348). (c) Bone marrow or stem cell transplant (860).
    }
    \label{fig:heatmap}
\vspace{-0.3cm}
\end{figure*}

In the following analysis, we examined the top-scored EHR codes under high-risk subphenotypes across the four modalities for three high-risk phenotypes to assess the impact of subphenotypes on disease severity.

\paragraph{Cirrhosis of liver without alcohol subphenotypes}
We focused on Cirrhosis due to its high mortality rate, ranging from 34\% to 69\%~\cite{da2021assessing}. The subphenotype 571.51-2 represents the most advanced stage, marked by severe complications and high mortality risk. The ICD modality (Fig.\ref{fig:heatmap} a, ICD Modality) highlights severe cases such as  "Autoimmune hepatitis (571.42)", which is associated with jaundice and amenorrhea ~\cite{manns2001autoimmune}. In contrast, subphenotype 571.51-0 includes conditions like chronic hepatitis C with hepatic coma (070.44) and hepatorenal syndrome (572.4), often managed with antiviral therapies ~\cite{hepatitisCprogression}. Subphenotype 571.51-1 is characterized by features such as esophageal varices without bleeding (456.1) and unspecified viral hepatitis C with hepatic coma (070.71).

In the medication modality, methylprednisolone is a high-probability medication for subphenotype 571.51-2, used to treat severe inflammation and immune response complications ~\cite{methylprednisolone}. In the CPT modality, it shows that critical procedures such as cholecystectomy is associated with high morbidity ~\cite{cholecystectomy,biliarytractoperations}. Doctor notes frequently mention 'cirrhosis', 'lactulose', and 'encephalopathy' for subphenotype 571.51-2, highlighting its severity ~\cite{lactulose, HEcomplications} (Fig.\ref{fig:heatmap} a).

\paragraph{Brain injury subphenotypes}
We observed that a subphenotype related to brain condition exhibits a high association with mortality (Fig.\ref{fig_SHAP_value}). PheCode 348 (Other conditions of brain) is stratified into subphenotypes, with subphenotype 348-2 being the most severe, including conditions such as brain death (348.82) and patients receiving palliative care (V66.7) ~\cite{starr2024brain, goni2023makes}. In contrast, subphenotype 348-0 includes conditions like central nervous system complications (997.01) and unspecified persistent mental disorders (294.9), managed with supportive care. Subphenotype 348-1 includes anoxic brain damage (348.1), indicating potential for varying outcomes ~\cite{lacerte2023hypoxic}. Therefore, subphenotype 348-2 is the most severe subphenotype due to the inclusion of brain death and palliative care, indicating end-of-life conditions (Fig.\ref{fig:heatmap} b).

\paragraph{Bone marrow transplant subphenotypes}
Subphenotypes 860-0 and 860-2 for Bone Marrow or Stem Cell Transplant are both highly associated with mortality (Fig.\ref{fig_SHAP_value}). Subphenotype 860-0 represents a slightly more severe stage than 860-2. The ICD modality indicates that subphenotype 860-2 involves high-probability conditions such as complications of bone marrow transplant (996.85) and peripheral stem cells replaced by transplant (V42.82), highlighting its severity. In comparison, subphenotype 860-0 includes conditions like adrenal cortical steroids causing adverse effects in therapeutic use (E932.0), family history of malignant neoplasm of the gastrointestinal tract (V160), and multiple myeloma in relapse (203.02) (Fig.\ref{fig:heatmap} c, ICD Modality).

Based on the medication modality (Fig.\ref{fig:heatmap} c), high-probability medications for subphenotype 860-2 include acyclovir, ursodiol, and atovaquone. Acyclovir reduces the probability and delays the onset of cytomegalovirus infection ~\cite{acyclovir}. Ursodiol prophylaxis decreases hepatic complications after allogeneic bone marrow transplantation ~\cite{ursodiol}. Atovaquone is crucial for anti-Pneumocystis prophylaxis post-transplant ~\cite{atovaquone}. In contrast, subphenotype 860-0 is associated with medications like loperamide for chemotherapy-induced diarrhea ~\cite{loperamide}, clotrimazole for fungal infections, and cyclosporine to prevent organ rejection ~\cite{cyclosporine}.

The CPT modality indicates that subphenotype 860-2 requires critical procedures such as the injection or infusion of cancer chemotherapeutic substances and biological response modifiers (BRM), which are associated with severe complications. For subphenotype 860-0, CPT codes include allogeneic hematopoietic stem cell transplant without purging, transjugular liver biopsy, and implantation of chemotherapeutic agents, indicating complex procedures (Fig.\ref{fig:heatmap} c, CPT Modality). Additionally, the doctor notes for 860-2 often mention 'BMT' (bone marrow transplant), 'GVHD' (graft-versus-host disease), and 'TPN' (total parenteral nutrition) as high-probability features. For 860-0, the doctor notes frequently mention 'cyclosporin' and 'caspofungin', indicating the use of immunosuppressants and antifungals to manage post-transplant complications (Fig.\ref{fig:heatmap} c, Note Modality). Although both 860-0 and 860-2 are severe subphenotypes, 860-0 is slightly more severe due to its advanced stage, requiring more intensive medication management and higher-risk medical interventions

\subsection{Insulin usage prediction in PopHR}
\begin{figure*}[t!]
\centering
\includegraphics[width = 0.9\linewidth]{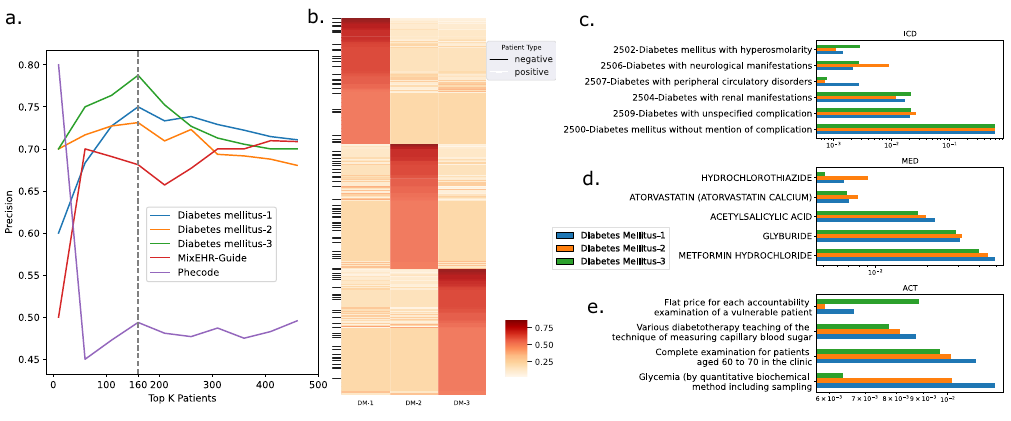}
\vspace{-0.6cm}
\caption{
  \textbf{Predicting future insulin usage among diabetic patients based on their subphenotype risks.}
  (a) Precision curve of top K patients ranked by risk $(\uptheta)$ in each subphenotypes and Diabetes PheCode. We used each phenotype score as the prediction score for the future insulin usage without training any supervised classifier. The AUROC for each phenotype scores are displayed in the legend.
  (b)  Patient-diabetes topic mixture $(\uptheta)$ of top 160 patients from each of the 3 diabetic subphenotypes. Each row represents a unique patient, that is, there is no overlap between the top 160 patients among the three subphenotypes. We chose the top 160 patients where the precision peaks for all three subphenotypes.
  (c-e) Top ICD, medication, and ACT codes for the 3 diabetes subphenotypes. For each subphenotype, we selected the top 4 codes with the highest subtopic probabilities and displayed the union of the top codes across the 3 subphenotypes.  
}
\label{fig6:Diabetes-insulin prediction}
\vspace{-0.3cm}
\end{figure*}

For the top risked patients in each subphenotype, we observed differences in their predictive power.Diabetes mellitus-3 demonstrated the highest precision up to the top 250 patients, while Diabetes mellitus-1 surpassed Diabetes mellitus-2 at around top 130 patients and Diabetes mellitus-3 at around 250 patients. All three subphenotyeps exhibited equal or higher precision compared to MixEHR-G and significantly better precision than using raw phecodes (Fig.\ref{fig6:Diabetes-insulin prediction} a). We visualized the patient-diabetes topic mixture $\uptheta$ for the top 160 patients to assess the proportion of positive patients and the distribution of $\uptheta$ among the top patients in each subphenotype (Fig.\ref{fig6:Diabetes-insulin prediction} b), where each row represents a unique patient. 

The three subphenotypes were differentiated by their top ICD-9 codes, prescription, and ACT codes (Fig.\ref{fig6:Diabetes-insulin prediction} c-e). First, Diabetes mellitus-1 likely represents a diabetic condition with multiple complications, especially cardiovascular, in older patients. This is indicated by its highest probability with peripheral circulatory disorder and the relative lower probabilities with renal, neurological and hyperosmolar manifestations. For Diabetes mellitus-1, the highest probability in acetylsalicylic acid and metformin, along with moderate probability with other drugs (hydroclorothizae, atorvarstatin and glyburide) suggests the importance of controlling blood sugar and cardiovascular risks ~\cite{doi:10.1056/NEJMoa1804988}. Diabetes mellitus-1  requires regular blood sugar check-ups and comprehensive clinic examinations. Diabetes mellitus-2 is associated with the highest likelihood of diabetes with unspecified complications as well as neurological manifestations. Although its most probable medications hydrochlorothiazide and atorvastatin are primarily used to treat hypertension and manage cholesterol levels, there have been reports of neuropathy risks associated with these drugs. Diabetes mellitus-3 appears as a severe condition with significant renal and hyperosmolar manifestations. Symptoms often emerge in later stages where regular drug treatments (metformin, glyburide) are not as effective, necessitating  immediate medical treatment. This is evident from Diabetes mellitus-3's highest probability with vulnerable patient and lowest probability with all types of drugs shown here.

\subsection{Age prevalence of subphenotypes in PopHR}
\begin{figure}[t!]
\centering
\includegraphics[width = \linewidth]{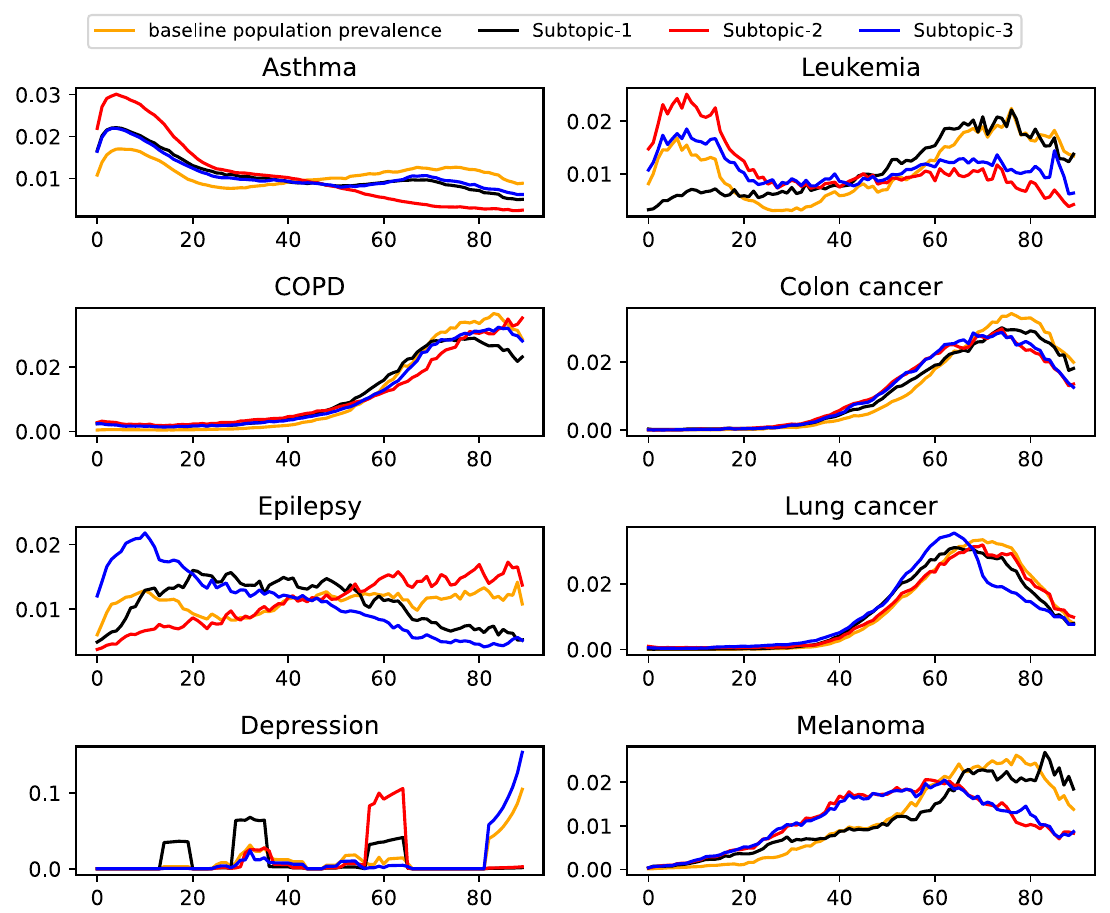}
\vspace{-0.6cm}
\caption{
  \textbf{Estimate of relative phenotype prevalence.}
Based on the patient-topic mixtures $\uptheta$, we computed relative phenotype prevalence stratified by age for 8 diverse disease phenotypes normalized by the mean estimate over time (Methods). We compared the predicted subtopic prevalences using \model with the baseline prevalence estimate based on raw phecode frequency.}
\label{fig7:Prevalence of 8 diseases}
\vspace{-0.3cm}
\end{figure}

We leveraged the inferred patient topic mixtures $\uptheta$ to investigate the population-level prevalence of  subtopic for 8 diverse phenotypes with respect to age and compared with the baseline using the PheCode counts (Fig.\ref{fig7:Prevalence of 8 diseases}).
These diseases were selected as a proof-of-concept due to their diverse onsets across ages. Notably, the subphenotype trends exhibit more distinct patterns compared to the baseline trend, which appears averaged over the 3 subphenotype trends. For instance, asthma is known to exhibit two age peaks at 10 and 75. The inferred subphenotype asthma-2 is more prevalent at an earlier age, but less so at the later in life. In contrast, the baseline prevalence exhibits a much flatter pattern, averaging out the salient features from the three subphenotype prevalence trends. 
Leukemia exhibits two modes: one at around 10 years old as shown by leukemia-2 and leukemia-3; another  at around 75 years old as shown by leukemia-1. 
Similarly, the three epilepsy subphenotypes show three stages of clear prevalence, which is unobserved in baseline method. Specifically, epilepsy-3 is more prevalent at early age; epilepsy-1 peaks at middle ages; and epilepsy-2 continues to rise into senior age. 
Depression subphenotype also exhibits intriguing pattern, indicating adolescence (depression-1), average age of pregnancy (depression-1), and menopause (depression-2). Depression-3 resembles the baseline population prevalence,  remaining relatively low and stable across all ages, except after. Melanoma-2 and -3 exhibit similar prevalence patterns with a peak at 60 years old, while malanoma-1 and baseline show a delayed but significant rise in prevalence at a later age. In contrast, the prevalence for COPD, colon cancer and lung cancer is rather consistent among subphenotypes and baseline, with peaks at 80, 75, and 65 years old, respectively. 

\section{Discussion}
The increasing adoption of EHRs and the advancement of automatic subphenotyping techniques provide numerous opportunities for healthcare applications such as personalized risk prediction and precision medicine. This stemmed from the realization that patients with seemingly homogeneous diseases often display varying severity and require potentially drastically different treatment plans according to the underlying cause of the symptoms. 
In this study, we propose a nested-guided topic modeling algorithm called \model to simultaneously model the subtopics for more than 1500 phenotypes based on two high-dimensional and multimodal population-level EHR datasets. \model can be guided by any expert-curated phenotype codes, nested with several subtopics controlled as a hyperparameter. MixEHR-Nest is highly scalable because of the use of the bag-of-words representations and the efficient collapsed Gibb-sampling topic inference algorithm inherited from its predecessors ~\cite{griffiths2004finding}.
The learned latent subtopic mixtures are identifiable for all the diseases because we can confidently observe how the subtopics are associated with its main phenotype prior and understand the composition of each subtopic. In proof-of-concept analyses, \model's learned subtopic mixtures yields better performance in both mortality prediction and insulin usage prediction. We observe that the mortality risks of different subtopics are well-aligned with their learned mixtures, and that the relative population prevalence of sub-phenotypes are clearly indicative of different the known ages of onsets for diseases like asthma, leukemia, epilepsy, and depression, which are helpful in monitorying population health.

The inherent complexity of disease mechanisms and the diversity of patient profiles present significant challenges to subphenotyping, requiring an incorporation of  expert knowledge into subphenotype topic inference using phenotype concepts from PheWAS and CCS mapping. However, it remains challenging to determine the optimal number of subtopics for single diseases, such as acute respiratory distress syndrome (ARDS) ~\cite{maddali2022subtype, calfee2014subtype, Sinha2018subtype} and sepsis ~\cite{seymour2019derivation, mayhew2018flexible, bhavani2019identifying}. To solve this issue, \model provides a quantitative, instead of empirical, estimate to the subphentyping potentials in multiple diseases. By assessing the disparity between subphenotypes, we understand the subphenotype topics with each disease, which can be improved using complementary information from other modalities.
Moreover, characterizing complex traits requires the insights from genomics, epidemiology, and clinical practice. For example, leveraging genomic data, particularly through the use of genome-wide association studies (GWAS), enables a transition from phenotype to genotype, allowing for the identification of biomarkers for subphenotyping. 
Together, the interdisciplinary approach that incorporates genomics insights with epidemiological, and expert knowledge about disease sub-types may help us overcome current barriers, leading to more accurate and effective strategies for personalized medicine.


\bibliographystyle{ACM-Reference-Format}
\bibliography{main}




\onecolumn
\appendix

\section{Appendix}
\subsection{An example of prior initialization}\label{sec:init_prior_eg}
In the case of MIMIC-III,f patient $d$ has PheCode $k$, we raise the values in the $k^{th}$ row of $\bm{\upalpha}_d$ to relatively high value (e.g., 0.9) compared to the hyperparameters corresponding to the unobserved PheCode (e.g., values randomly sampled from a range of [0.001, 0.01]). For example, suppose we have $K=4$ PheCode-guided topics and $M=3$ sub-topics per topic. If the patient $d$ has ICD-9 code that belongs to the definition of PheCode 2 but nothing else, then the topic prior hyperparameter matrix $\bm{\upalpha}_d$ is set to:
\[
    \bm{\upalpha}_d =\begin{bmatrix}
        0.005 & 0.008 & 0.002\\
        0.9 & 0.9 & 0.9\\
        0.003 & 0.002 & 0.002\\
        0.006 & 0.007 & 0.005\\
    \end{bmatrix}
\]
In the case of PopHR,  we raise the values in the $k^{th}$ row of $\bm{\upalpha}_d$ to the two-component Poisson and Lognorm mixture model estimate, while keeping other rows 0.

\subsection{Example of initialization of sufficient statistics}\label{sec:eg_init_suff_stat}
For example, suppose a patient $d$ has two ICD tokens $x^{(\text{ICD})}_{1d}$, $x^{(\text{ICD})}_{2d}$ and one non-ICD token $x^{(\text{non-ICD})}_{1d}$. The ICD token $x^{(\text{ICD})}_{1d}$ has been assigned with one of the three subtopics $k_1, k_2, k_3$ and $x^{(\text{ICD})}_{2d}$ with one of the three subtopics $k'_1, k'_2, k'_3$. For token $x^{(\text{non-ICD})}_{1d}$, we randomly assign it with with one of the six subtopics $k_1, k_2, k_3, k'_1, k'_2, k'_3$. We then increment $n^{(\text{non-ICD})}_{w \cdot k_m}$ by 1 and $n_{\cdot dk_m}$ by $\eta$ (cf. above) to mitigate the potentially inaccurate topic assignment in non-ICD modality.  

\subsection{LDA using collapsed Gibbs Sampling Derivation}\label{sec:lda_derive}
We describe the key derivations of the topic inference for the basic LDA ~\cite{griffiths2004finding}. 

\paragraph{Collapsed Gibbs sampling of topic assignments} 
We first integrate out the Dirichlet variables $\uptheta$ and $\upphi$:
\begin{align}
    & p\left(z\vert\alpha\right)=\int p\left(\uptheta\vert\alpha\right)p\left(z\vert\uptheta\right)d\uptheta=\prod_d\frac{\Gamma(\sum_{k}\alpha_{dk})}{\prod_{k}\Gamma(\alpha_{dk})}\frac{\prod_{k}\Gamma(\alpha_{dk}+n_{.dk})}{\Gamma(\sum_{k}\alpha_{dk}+n_{.dk})} \nonumber \\
    & p\left(x|z\right)=\int p\left(\upphi |\beta \right)p\left(x|z,\upphi \right)d\upphi =\prod_{k}\frac{\Gamma \left(W\beta \right)}{\prod _{w}\Gamma \left(\beta \right)}
    \frac{\prod^{W}_{w=1}\Gamma \left(\beta +{n}_{w.k}\right)}{\Gamma (W\beta +\sum_{w}{n}_{w.k})}
\end{align}

The conditional probability of topic assignment $z_{id}$ for token $i$ in document $d$ given the topic assignments for the rest of the $N_d - 1$ tokens $z^{-(id)}$ has a closed form expression:
\begin{align}
    &p\left(z_{id}=k \vert{x_{id}=w,z}^{-\left(id\right)},x^{-\left(id\right)}\right) \nonumber \\
    & \propto p\left(z_{id}=k,z^{-\left(id\right)}\vert\alpha_{dk}\right)p\left(x_{id}=w,x^{-\left(id\right)}\vert z_{id}=k,z^{-\left(id\right)}\right) \nonumber \\
    &\propto\prod_{k'\neq k}\Gamma\left(\alpha_{dk'}+n_{.dk'}^{-\left(id\right)}\right)\Gamma\left(\alpha_{dk}+n_{.dk}^{-\left(id\right)}+1\right) \nonumber \\
    &\prod_{k'\neq k}\frac{\prod_w\Gamma\left(\beta+n_{w.k'}^{-\left(id\right)}\right)}{\Gamma(W\beta+\sum_wn_{w.k'}^{-\left(id\right)})}\frac{\Gamma\left(\beta+n_{x_{id}.k}^{-\left(id\right)}+1\right)}{\Gamma(W\beta+\sum_wn_{w.k}^{-\left(id\right)}+1)} \nonumber \\
    &\propto\left(\alpha_{dk}+n_{.dk}^{-\left(id\right)}\right)\left(\frac{\beta+n_{x_{id}.k}^{-\left(id\right)}}{W\beta+\sum_wn_{w.k}^{-\left(id\right)}}\right)
\end{align}

We perform Gibbs sampling to assign new topics for each token in each document, where $n^{-\left(id\right)}_{\cdot dk}$ is the count of tokens in current patient $d$ assigned to subtopic $k$ without counting the current $i$th token; $n^{-\left(id\right)}_{w\cdot k}$ is the total counts of the EHR code $w$ across all the patients without counting the current $i$-th token in current patient $d$:
\begin{align}
    & z_{id} \vert z^{-(id)} \sim \prod_{k=1}^{K} 
    \left(\alpha_{dk}+n_{\cdot dk}^{-\left(id\right)}\right) \left(\frac{\beta+n_{x_{id}\cdot k}^{-(id)}}{W\beta + \sum_{w=1}^W n_{w\cdot k}^{-(id)}}\right) \\
    & \text{where  } n_{.dk}^{-(id)} = \sum_{i'\neq i}^{N_d}{\lbrack z}_{i'd}=k\rbrack \nonumber \\
    & n_{w.k}^{-\left(id\right)} = \sum_{d'\neq d\;\mathrm{or}\;i'\neq i}\left[z_{i'd'}=k,x_{i'd'}=w\right]
\end{align} 

Upon iteratively sampling the topics for all tokens and all documents, the expected topic mixture and topic distribution are: 
\begin{align}
    \hat{\uptheta}_{dk} &\equiv \mathbb{E}\left[\uptheta_{dk} \mid z_{d}\right] 
    = \frac{\alpha_{dk} + n_{\cdot d,k}}{\sum^{K}_{k=1}\alpha_{dk} +  n_{\cdot dk}} \nonumber \\
    \hat{\upphi}_{wk} &\equiv \mathbb{E}\left[\upphi_{wk} \mid z_{k}\right]=\frac{\beta+n_{w \cdot k}}{W \beta+\sum^{W}_{w^{\prime}=1} n_{w^{\prime} \cdot k}}
\end{align}

\subsection{Non-ICD topic inference}\label{sec:non_icd_topic_infer}
\begin{align}
    & z^{(\text{non-ICD})}_{i'd}\sim \prod_{k_m=1}^{K\times M} 
    \left(\alpha_{dk_m} + n_{\cdot dk_m}^{-\left(i', d\right)}\right) \left(\frac{\beta+n_{x_{i' d} \cdot k_m}^{-(i', d)}}{ W_{\text{non-ICD}}\beta+\sum_{w=1}^{W_{\text{non-ICD}}}n_{w \cdot k_m}^{-(i', d)}}\right)\\
    & n_{.dk_m}^{(\text{non-ICD})} = \sum_{i'=1}^{N_d^{(\text{non-ICD})}} [z^{(\text{non-ICD})}_{i'd} = k_m] \nonumber \\
    & n_{\cdot dk_m} = \hat{n}_{.dk_m}^{(\text{ICD})} + \eta \times n_{.dk_m}^{(\text{non-ICD})}\label{eq:n_dk} \nonumber \\
    & n^{(\text{non-ICD})}_{w\cdot k_m} = \sum_{d=1}^D \sum_{i'=1}^{N_d^{(\text{non-ICD})}} [z^{(\text{non-ICD})}_{i'd}=k_m][x^{(\text{non-ICD})}_{i'd}=w] \nonumber\\
    & \text{for}\ \ w \in \ \{ 1, \ldots, W_{\text{(non-ICD)}}\}    
\end{align}

\subsection{Longitudinal topic inference}\label{sec:longitudinal_topic_infer}
Specifically, the expected topic mixture $\hat{\uptheta}_d$ across all modalities is:
\begin{align}    
    & \hat{\uptheta}_{dk_m} \equiv \mathbb{E}\left[\uptheta_{dk_m} \mid z_{d}\right] = \frac{\alpha + n_{\cdot d k_m}}{\sum^{K\times M}_{k_m'=1}\alpha_{dk'_m} +  n_{\cdot d k_m'}} \nonumber \\
    & \text{where} \quad n_{\cdot d k_m} = \sum^{T}_{t =1}n^{(t)}_{\cdot d k_m}\label{eq_11}
\end{align}
and the log likelihood is:
\begin{align}
    & \mathcal{L} =\sum^{T}_{t=1}\sum^D_{d=1} \sum^{N^{(t)}_{d}}_{i=1}\sum^{K\times M}_{k_m=1}\log\hat{\uptheta}_{dk_m} \hat{\upphi}^{(t)}_{x^{(t)}_{id}k_m}\label{eq_12}
\end{align}

\renewcommand{\thefigure}{S\arabic{figure}}
\setcounter{figure}{0}
\subsection{Supplementary Figure}

\begin{figure*}[!t]
\centering
\includegraphics[width =0.8\textwidth]{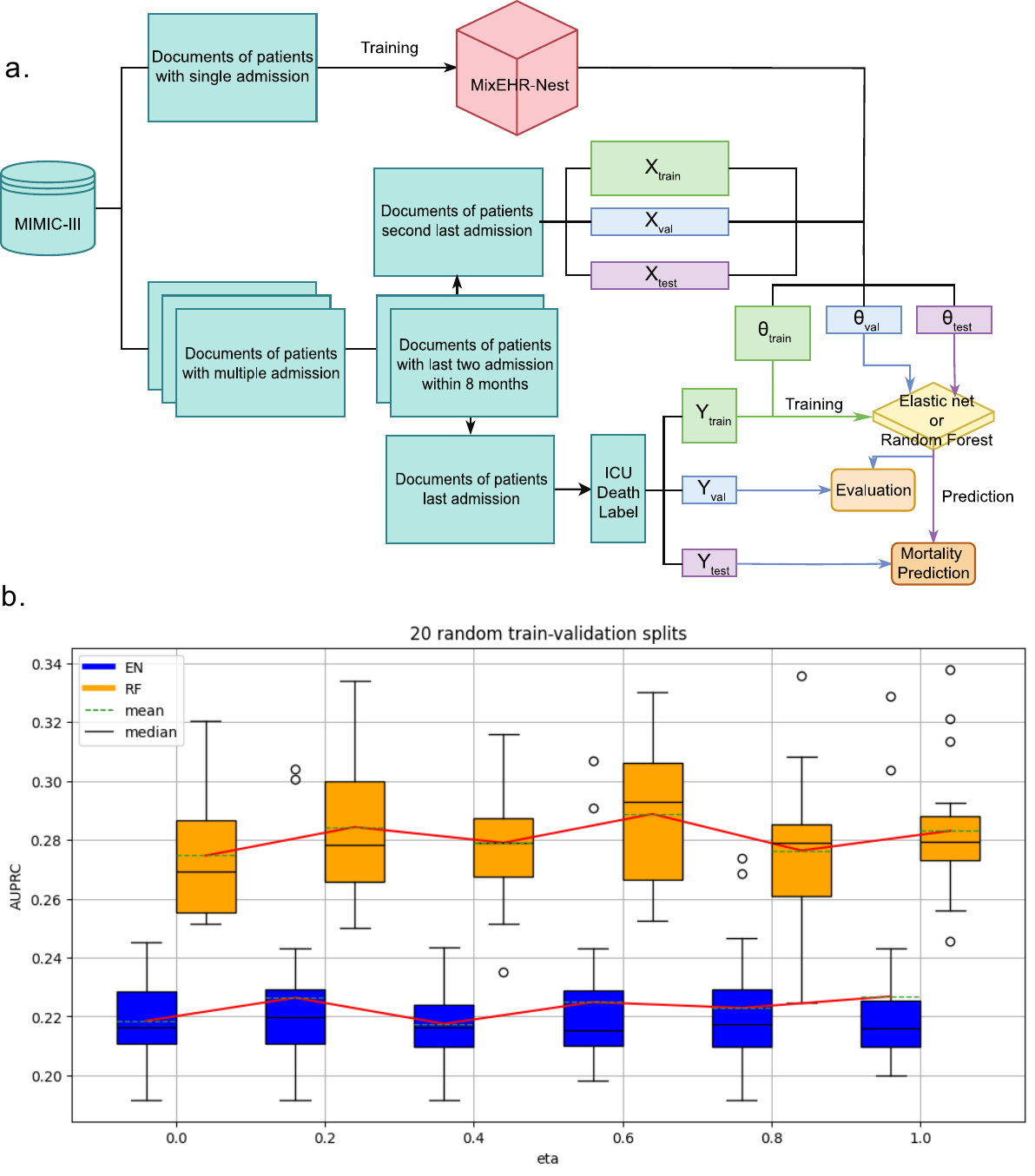}
\caption{\textbf{Schematic diagram of the $\eta$ selection workflow.} (a) The MIMIC-III data, after preprocessing, is first divided into patients with single admissions and multiple admissions. The documents of patients with single admissions are used to train the model with different $\eta$ values. Patients with multiple admissions are further filtered by the condition that the time difference between their last two admissions is less than 8 months. The documents from these selected patients' second-to-last admissions are then split into a training set ($X_{train}$), test set ($X_{test}$), and validation set ($X_{val}$). The mortality status at their last admission is used as labels ($Y_{train}$, $Y_{test}$, and $Y_{val}$). The model trained on single-admission patients' documents is used to infer $\uptheta_{train}$, $\uptheta_{test}$, and $\uptheta_{val}$ for the training, test, and validation sets, respectively. The $\uptheta_{train}$ along with $Y_{train}$ are input into a Random Forest Classifier or elastic net, and the trained model is then used to make predictions for $\uptheta_{val}$. The $\eta$ with the highest mean AUPRC of prediction 20 random validation sets is selected as the optimal $\eta$. Finally, mortality is predicted on the test data using $\uptheta_{test}$ from the model with the selected $\eta$. (b) Boxplot showing performance of different combinations of $\eta$s and elastic net or random forest classifier on 20 random train-validation splits. 
}
\label{sup-2}
\end{figure*}

\begin{figure}[!t]
\centering
\includegraphics[width = 0.5\textwidth]{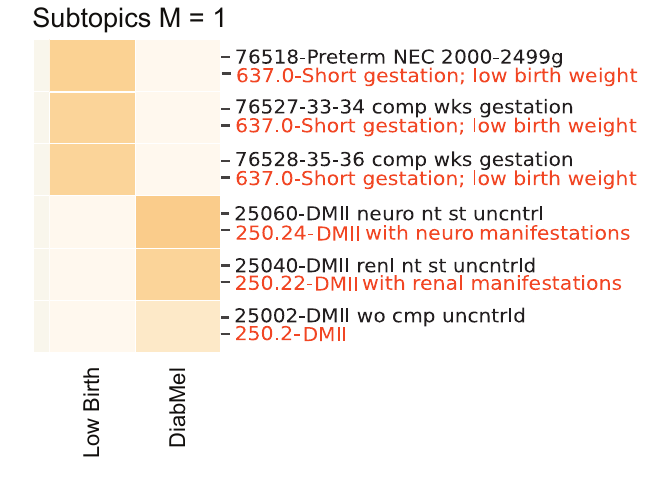}
\caption{
Top ICD codes inferred by \model for target phenotype topics, guided by CCS codes. No subtopic within a single phenotype.
}
\label{sup-1}
\end{figure}

\begin{figure}[t!]
\centering
\includegraphics[width = 0.5\linewidth]{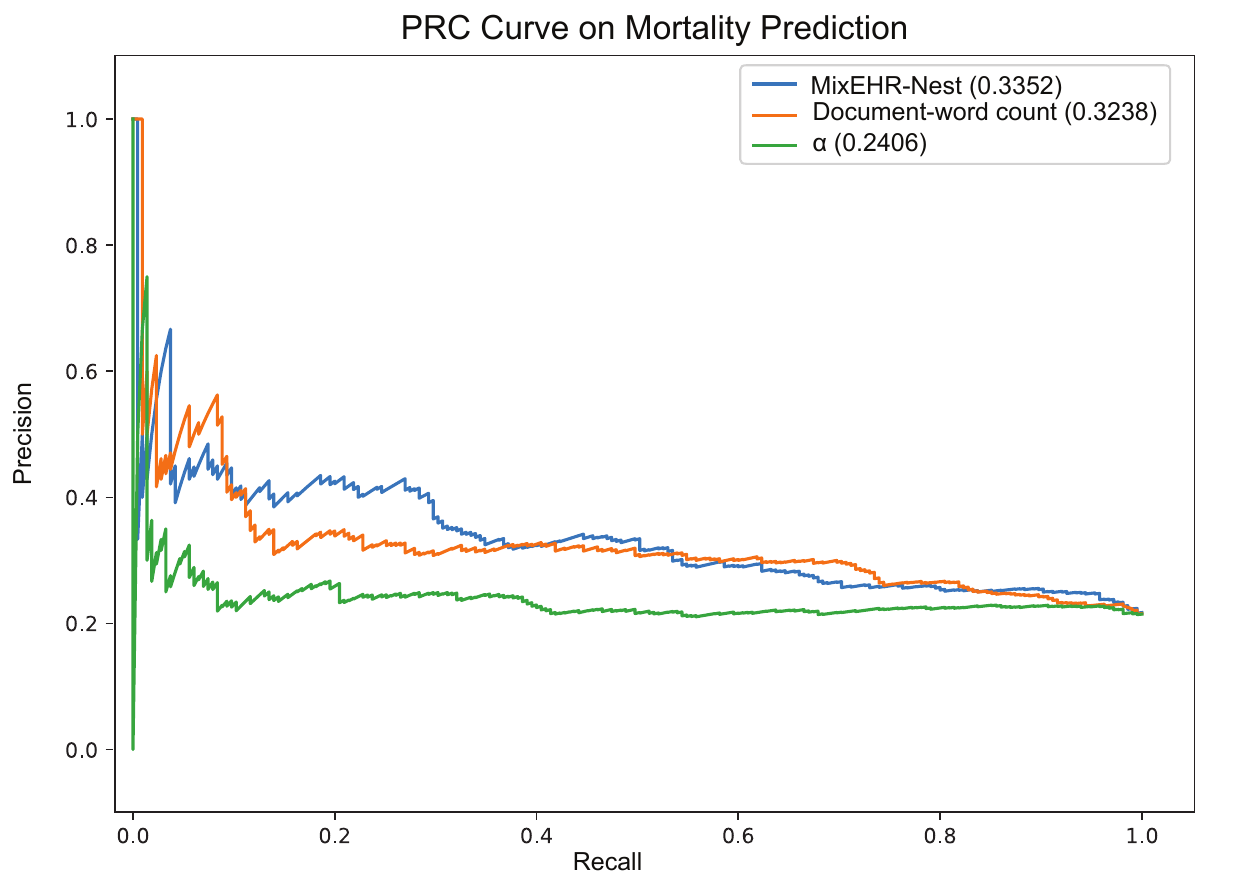}
\caption{Additional analysis benchmarking model performance on mortality prediction for multiple admissions, by precision and recall(AUPRC).}
\label{sup_prc}
\end{figure}

\begin{figure*}[t!]
\centering
\includegraphics[width = 0.6\linewidth]{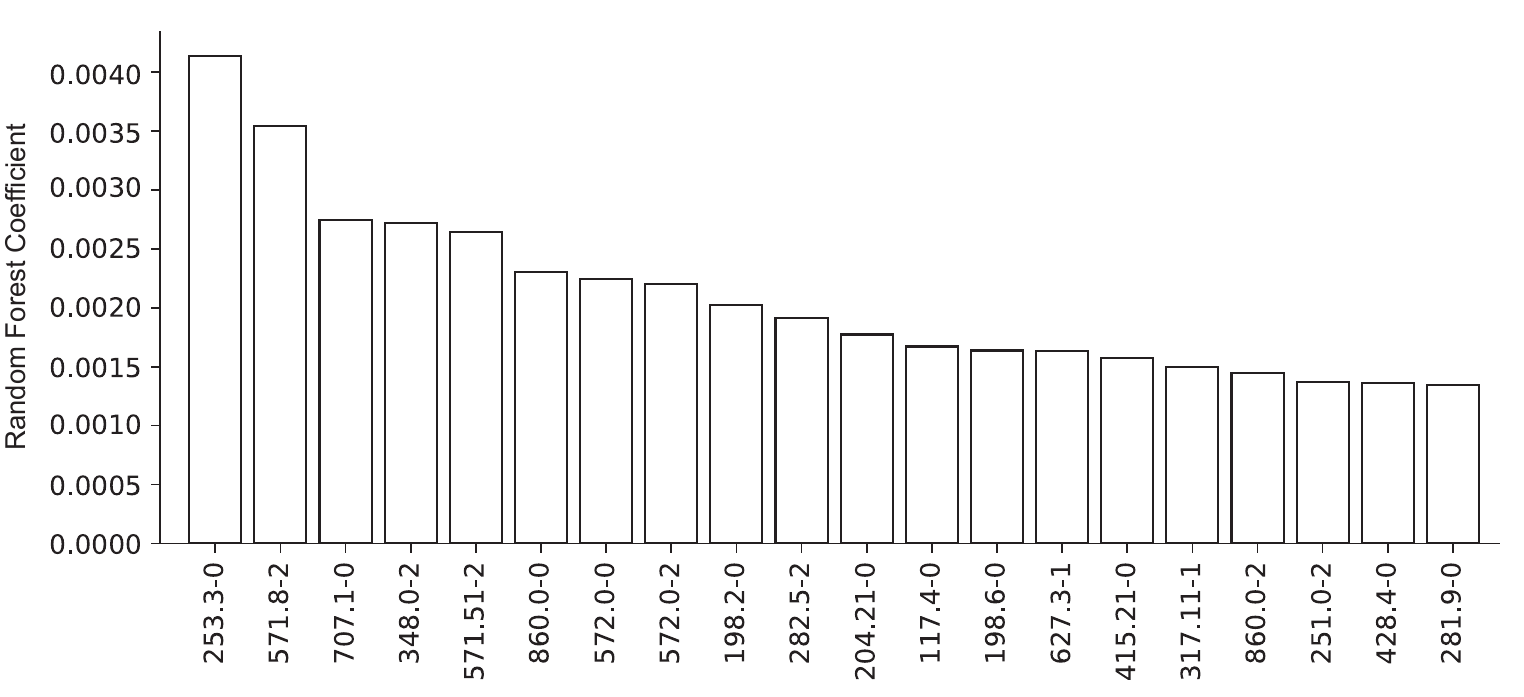}
\caption{
  \textbf{Additional analysis of high-risk mortality disease predicted by \model.} Top 20 disease subphenotypes with the highest mortality risk, as selected by Random Forest feature importance.
  }
\label{sup_SHAP_value}
\end{figure*}

\end{document}